\documentclass[runningheads]{llncs}
\usepackage{graphicx}
\usepackage{amsmath}
\usepackage{subcaption}
\usepackage{hyperref}

\usepackage{color}

\begin{document}
%
\title{Benchmarking Tropical Cyclone Rapid Intensification with Satellite Images and Attention-based Deep Models}
\titlerunning{Benchmarking TC RI with Satellite Images and Deep Models}
%
\author{Ching-Yuan Bai\inst{1} \and
Buo-Fu Chen\inst{2}\and
Hsuan-Tien Lin\inst{1}}
\authorrunning{C. Y. Bai et al.}
%
\institute{Department of Computer Science and Information Engineering, \\
National Taiwan University, Taipei, Taiwan \\
\email{\{b05502055@ntu.edu.tw, htlin@csie.ntu.edu.tw\}} \and
Center for Weather Climate and Disaster Research, \\
National Taiwan University, Taipei, Taiwan\\
\email{bfchen777@gmail.com}}
\maketitle              
\setcounter{footnote}{0}    
\begin{abstract}
Rapid intensification (RI) of tropical cyclones often causes major destruction
to human civilization due to short response time. 
It is an important yet challenging task to accurately predict this kind of extreme weather event
in advance.
Traditionally, meteorologists tackle the task with human-driven
feature extraction and predictor correction procedures.
Nevertheless, these procedures do not leverage the power of
modern machine learning models and abundant sensor data, such as satellite
images. 
In addition, the human-driven nature of such an approach makes it difficult to reproduce and benchmark
prediction models. 
In this study, we build a benchmark for RI prediction using only satellite
images, which are underutilized in traditional techniques. 
The benchmark follows conventional data science practices, making it easier for data scientists to contribute to RI prediction.
We demonstrate the usefulness of the benchmark by designing a domain-inspired
spatiotemporal deep learning model. 
The results showcase the promising performance of deep learning in solving
complex meteorological problems such as RI prediction.

\end{abstract}
\section{Introduction}
\label{sec:introduction}
The tropical cyclone (TC) is one of the most devastating weather systems on Earth,
characterized by intense and rapidly rotating winds around a low-pressure
center and associated with eyewall clouds and spiral rainbands producing heavy
rainfall. 
In order to reduce and respond to damage caused by TCs, the past half-century
has seen much effort devoted to improving the forecast of TC
track, intensity, and the associated rainfall and flooding.
Although TC track forecasts has improved significantly during the past
few decades, prediction of TC rapid intensification (RI) remains challenging,
which affects the subsequent production of TC structure and rainfall
forecast~\cite{gall2013hurricane}. 
TC intensity is defined as the maximum sustained wind in the TC inner-core
region, and rapid intensification (RI) is defined as a TC experiencing an
intensity increase surpassing a threshold (25--35 knots) within a 24-hour period.
The range of thresholds represents the 90 to 95 percentiles of 24-hour TC intensity
changes in different basins.

Accurately predicting the onset of RI is particularly crucial because reacting
to an off-shore RI event before TC landfall requires sufficient time;
delayed reaction has caused some of the most catastrophic TC disasters.
For instance, hurricane Harvey in August 2017 caused 107 casualties and cost
approximately \$125 billion USD as it rapidly intensified to a category~4
hurricane only hours before landfall.
However, improvement in RI prediction has been slow partly because RI events are rare.
Additionally, favorable environmental conditions are generally necessary but do not
guarantee RI onset~\cite{hendricks2010quantifying}.
Thermo-dynamical processes within the TC in response to these environmental
forces are believed to be even more critical. 
For example, the development of up-shear convective
burst~\cite{miyamoto2018structural,rogers2013airborne} and asymmetric surface
fluxes and boundary-layer inflow associated with the background
flow~\cite{chen2019LMF,rappin2012effect} are factors that control the rainfall
distribution and, in turn, TC intensification. 
Thus, a successful RI prediction scheme must accurately depict both
environmental conditions (in which a TC is embedded) and vortex-scale features
such as the distribution of precipitation or inner-core TC structure.

The goal of this paper is to tackle TC RI prediction from a 
data-science perspective.
We propose a new benchmarking procedure with 
rigorous practices common to data science
which includes a satellite image based dataset to be publicly
released.
We experiment with deep learning methods for this task.
In Section~\ref{sec:related_work}, we briefly review related work on TC
intensity and RI prediction.
In Section~\ref{sec:benchmark}, we introduce our newly devised benchmark and
the major improvements over previous benchmarks.
In Section~\ref{sec:proposed_method}, we propose an attention-based deep
learning approach for this task and highlight the connection between
model design and meteorological domain knowledge.
In Section~\ref{sec:experiment_and_analysis}, we present the experimental results
and attempt to interpret what the model has learned based on
attention mechanisms.
We conclude our findings in Section~\ref{sec:conclusion}.

\section{Related Work}
Previous studies on predicting TC RI focus on utilizing predictors as features (specifically predictors published by the SHIPS project, to be introduced in Section~\ref{sec:related_work_ships}).
Predictors are high level statistics (e.g., mean or standard deviation) that summarize collected or simulated atmospheric data in a span of time and space.
The high level physical meaning carried by predictors allow human to easily comprehend and design prediction models accordingly.
However, the major downside of studies relying on predictors is the lack of exploration for new techniques that better exploit the raw data.
The loss of detail in predictors has bottlenecked improvements in prediction performance.

In this work, instead of relying on predictors, we directly utilize raw data collected from satellite sensors with deep learning models.
Our model architecture is inspired by domain insights from the Advanced Dvorak Technique (ADT)~\cite{olander2005adt}, an operational technique for TC intensity forecasting based on raw data.
Details of ADT will be discussed in Section~\ref{sec:related_work_adt}.

\label{sec:related_work}
\subsection{Statistical Hurricane Intensification Predictive Scheme}
\label{sec:related_work_ships}
The SHIPS project has
developed a series of statistical models for probabilistic prediction of
RI based on atmospheric predictors~\cite{kaplan2010revised,rozoff2011new,rozoff2015improvements}. 
The SHIPS RI index (SHIPS-RII~\cite{kaplan2010revised}) predicts the
probability of a TC intensifying by at least 25, 30, and 35~kt within 24 hours.
This scheme uses simple linear discriminant analysis to determine the RI
probability based on a relatively small number ($< 10$) of predictors
describing mainly environmental factors and limited aspects of 
internal TC structure observed by meteorological satellites. 
Candidate predictors ($\sim 20$) for SHIPS-RII~\cite{kaplan2010revised} are
subjectively determined by human intelligence, and the final predictors used
for linear discriminant analysis are basin-dependent. 
The final model is not publicly available, but the details of model design are
well documented in publication and thus technically reproducible.

A subsequent work~\cite{rozoff2011new} uses Bayesian inference
(SHIPS-Bayesian) and logistic regression (SHIPS-logistic) to predict RI
probability.
The authors show that both SHIPS-Bayesian and SHIPS-logistic exhibit forecast
performance that generally exceeds the skill of SHIPS-RII; blending the three
models further improves performance. 
Another study~\cite{rozoff2015improvements} integrates an additional 4 to 6
predictors derived from satellite passive microwave (PMW) observations into the
SHIPS-logistic model and demonstrates a relative performance improvement from 53.5\% to 103.0\%
in the Atlantic compared to the original model.
More details will be discussed in Section~\ref{sec:benchmark_ships-rii}.
Note that these SHIPS-RII techniques are only applicable in the Atlantic and eastern Pacific basins. 
Thus, a new technique applicable for all global TCs is demanded.

\subsection{Advanced Dvorak Technique}
\label{sec:related_work_adt}
ADT~\cite{olander2005adt} is an automated technique for real-time TC intensity
estimation based on satellite image analysis.
The technique replaces several human steps in the Dvorak Technique (DT) with automated procedures.
The ADT has two stages: scene type classification and intensity
post-processing.
A satellite image instance is first classified into one of many
pre-determined scene types according to the cloud distribution with respect to
the TC center.
Scene scores are derived from the characteristic matching test for each scene
type, and the image is classified as the scene type with the highest
corresponding scene score.
For instance, the curved band cloud scene characteristic matching test 
measures how well the curve of the cloud bands matches the $10^{\circ}$ log
spiral.
For each scene type, there is a unique method of deriving the intensity of the
satellite image.
In the second stage, the predicted intensity is post-processed by applying 
heuristic rules to clip the intensity into a reasonable range, 
after which the value is smoothed by applying a weighted average over the intensity of 
previous time steps.
Although ADT is currently used only to estimate TC intensity, it inspires
our proposed model in (a) focusing on individual frames to relate cloud features and TC intensity in the first stage; (b) averaging to take the time information into account in the second stage.
More details are presented in 
Section~\ref{sec:proposed_method}.

\section{Benchmark}
\label{sec:benchmark}
A fair and reproducible benchmark is the key to validating model performance for the continuous improvement of RI prediction. 
This includes adequately chosen metrics to reflect model performance and
well-defined, disjoint training and testing datasets to evaluate how well the
model generalizes on holdout data.
In this section, we first review the limitations of the current benchmark within the development of SHIPS-RII. Then, we propose a revised benchmark that solves the limitations and is better aligned with data science practices.


\subsection{Existing Benchmark within SHIPS-RII Development}
\label{sec:benchmark_ships-rii}

The metric adopted in the meteorology domain for benchmarking RI prediction is the Brier score
(BS), which is the mean square error (MSE) between
the $\{0, 1\}$-valued ground-truth RI labels and the probabilistic ($[0, 1]$-valued) predictions.
Leave-one-year-out cross validation of TC data from 1995 to 2013 is applied for
model training and hyperparameter tuning, with the Brier score serving as both the optimization objective and evaluation metric.
Cross-validated performance is reported with no special mention of holdout data
for testing.

The dataset for building SHIPS-RII is constructed as follows.
First, numerous features are collected from TCs, including climatology features
(statistical properties calculated from historical data), real-time measured
atmospheric/oceanic features, and features extracted from large-scale numerical
model simulations of current atmosphere conditions.
Then, a cherry-picked subset of TCs is removed for feature extraction to improve model performance.
Finally, a subset of summarized features is selected by linear discriminant
analysis and heuristics to serve as the final set of features (i.e. predictors in Section~\ref{sec:related_work_ships}).

Given the metric and dataset within the current benchmark above, we observe four main issues:
\begin{enumerate}
\item The original data for deriving the features is not easily accessible to the public. 
\item The cherry-picking data cleaning is subjective and not easily reproducible.
\item  There is no holdout test set to gauge the true generalization performance of models.
\item The BS metric is knowingly sensitive to class imbalance, but the RI prediction problem is class-imbalanced in nature.
\end{enumerate}

\subsection{Proposed Benchmark}
Next, we proposed a new benchmark to conquer the four issues. The new benchmark includes a more adequate evaluation metric and a publicly accessible dataset with a reproducible construction procedure and training/test splitting.


\subsubsection{Metric}
Different metrics are appropriate for evaluating different problems.
For instance, 
for problems that involve outliers in the data that one would like to ignore,
mean absolute error (MAE) is more suitable than mean square error
(MSE), which puts greater emphasis on outlying data points.
For the TC RI task, one major property of the data is its highly imbalanced
classes.
Due to the nature of the TC life cycle, RI mostly occurs only during the early to
mid stages of a TC when it grows in strength in a short period of time.
Thus, the number of RI to non-RI timeframes is approximately $1:20$ (the positive
class comprises $5\%$ of the total data).
Naturally, the performance metric should account for such data imbalance 
and should avoid rewarding highly biased classifiers that only output
the majority class.

Unfortunately, the Brier score does not take class imbalance into consideration.
For instance, a naive constant classifier that predicts only the majority class
(in this case, no-RI) with class imbalance of $1:20$ yields a very low Brier
score of $0.05$.
As we are more interested in discerning the minority class, it is difficult
for the Brier score to differentiate skillful models from unskillful models.

The Heidke Skill Score (HSS) is a metric commonly used in meteorology that accounts for class imbalance.
It is defined as how much better the model prediction accuracy is than the standard forecast:
$$
\text{HSS} = \frac{\text{ACC} - \text{SF}}{1 - \text{SF}}, \quad \text{ACC} = \frac{\text{TP} + \text{TN}}{N},\quad\text{and}
$$
$$
\text{SF} = \frac{\text{TP} + \text{FN}}{N} \times \frac{\text{TP} + \text{FP}}{N} + \frac{\text{TN} + \text{FN}}{N} \times \frac{\text{TN} + \text{FP}}{N},
$$
where $N$ is the number of total instances, $\text{ACC}$ is the accuracy,
$\text{SF}$ is the standard forecast, defined as correct by chance with class proportion, $\text{TP}$ is the number of true
positives, $\text{TN}$ is the number of true negatives, $\text{FP}$ is the
number of false positives, and $\text{FN}$ is the number of false negatives.
It is more difficult for the standard forecast to correctly predict under class imbalance as a skewed class proportion increases the odds of being correct by chance,
Thus, using standard forecast as the baseline makes HSS class-imbalance aware.

Other common metrics for handling imbalanced data include the F1 score (the harmonic
mean of precision and recall), the Matthews correlation coefficient,
the precision-recall area under the curve (PR-AUC), and the receiver operating
characteristic area under the curve (ROC-AUC) (the latter two are only for binary
classes).
We select PR-AUC as the main evaluation metric, since precision and recall are both important for this task,
coupled with HSS as a side metric to make it easier for meteorology experts to gauge performance.

\subsubsection{Dataset}
Since the raw form of most TC features is too complicated for people to
analyze and comprehend, only summarized statistical properties are released for
public usage and used by SHIPS-RII.
For example, one common feature is the standard deviation of 100--300~km
infrared brightness temperatures whereas the entire distribution of infrared
brightness temperatures over the whole geographical location is available but
not fully utilized.
Another common feature is the time-averaged potential TC intensity estimated by
simulation, even when data is available for the entire duration of the TC.
Information loss due to statistical properties taken over time or space is
particularly harmful for RI prediction because location- and timing-specific
properties of the tropical cyclone are known to cause rapid intensification.

Thus, we present a new dataset derived
from~\cite{chen2018rotation,chen2019estimating}, which consists of satellite 
images of tropical cyclones (four channels) with data available every three hours.%
\footnote{
Tropical Cyclone Rapid Intensification with Satellite Images:\\ \url{https://www.csie.ntu.edu.tw/~htlin/program/TCRISI/}
}
Each TC sequence is split into 24-hour segments.  Each segment is associated
with a label indicating the occurrence of RI during the period.
Preprocessing of satellite images follows typical meteorology practices with emphasis on the ``objectiveness'', i.e., the entire procedure involves no human intervention.
The data consists of TCs from 2003 to 2017 and is split into two parts for
training (2003 to 2014) and testing (2015 to 2017).
A disjoint testing set allows for the evaluation of model generalization on
holdout data; cross-validation is avoided because it is particularly
infeasible in terms of efficiency for deep learning models.

Satellite images, unlike features generated by predictive simulations, are
considered lower-level data, i.e., they involve less processing.
On the other hand, predictive features are generated by large-scale dynamic
atmospheric simulations.
Depending on predictive features poses a consistency issue because the
simulation models are updated yearly; thus they may change over a short period of time.
The fact that previous studies all depend on mixed data
underlines the importance of making better use of satellite images.

To highlight the contribution of our proposed benchmark, we inspect whether the issues with the existing benchmark are resolved.
First of all, the satellite image data including the scripts for preprocessing are released for public use.
The preprocessing procedure is fully automated and does not require any human intervention.
A disjoint set of data is reserved for testing to avoid presenting overfitting benchmark results.
Finally, PR-AUC and HSS are adopted as evaluation metrics, both better reflect model performance under class imbalance.


\section{Proposed Method}
\label{sec:proposed_method}
The new benchmark allows us to design and evaluate the potential of a domain-inspired model that takes in satellite images as inputs. The model belongs to the family of spatiotemporal deep learning methods, using deep
convolutional neural networks (CNNs) to extract essential features
from image data in an autonomous manner.
As models generally suffer from the curse of dimensionality,
complex spatiotemporal data requires manual feature extraction 
(such as hand-carved kernels) to compress it into smaller chunks.
Such manual processing discards a large amount of information and
is inefficient in finding optimal features because all feedback comes from
trial and error.
Currently the meteorology community is attempting to break through the
performance bottleneck by experimenting with modalities such as
lightning strike occurrence.

We opt for an orthogonal path by attempting to extract more from the data that we
already have, specifically TC satellite images with infrared (IR1), water
vapour (WV), visible light (VL), and passive microwave (PMW) channels.
Table~\ref{tab:feature_mapping} shows that satellite images contain
sufficient if not significantly more information in comparison to the
features adopted in SHIPS-RII.
One thing to note is the overlapping dependence on the infrared and water
vapor channels, with infrared strictly dominating water vapor.
Another is the lack of connection with the visible light channel as the signal
is useful for only half of the time (the other half is nighttime).
This is aligned with the feature selection in \cite{chen2018rotation},
which conclude that using only the IR1 and PMW channels yields the best performance. We follow the same practice to use only IR1 and PMW channels in our study.


\begin{table}
\centering
\caption{Potential satellite channels implicitly related to SHIPS-RII environmental conditions}
\label{tab:feature_mapping}
\begin{tabular}{|l|c|}
\hline
SHIPS-RII feature & Satellite channel \\
\hline
200-hPa divergence from 0--1000-km radius & IR1\\
850–700-hPa relative humidity from 200--800-km radius & IR1/WV\\
Total precipitable water averaged from 0--100-km radius & IR1/WV\\
Std dev of 50--200-km IR cloud-top brightness temperatures & IR1\\
\hline
\end{tabular}
\end{table}

\subsection{Model Architecture}
To recapitulate, the data for this task is spatiotemporal and highly 
class-imbalanced while relatively lacking in quantity, with only approximately 1900
tropical cyclone instances in total (from 2003 to 2017).
The constraint posed by data properties guides our model design.
As the tropical cyclone is a well-studied phenomenon in meteorology and many
techniques have been proposed and withstood the test of time, translating the
accumulated domain knowledge into data science is a difficult but rewarding task.
Specifically, the Advanced Dvorak Method (ADT) is the inspiration for the attention modules
in our model, which also allows us to observe what the model is focusing on.
More details regarding individual components of the model are discussed later
in this section.

\subsection{Base Structure}
Convolutional neural networks (CNNs) are suitable for modeling data with local
spatial dependency, as the convolution operator operates on neighboring pixels,
making it the base structure of our model.
We further extend the model to take into account the sequential nature of
data.
Currently there are three main neural network families that deal with time
series: 3D-CNN, recurrent neural networks (RNNs), and Transformer (from low to
high model complexity).
3D-CNN applies convolution on the time dimension in addition to the two spatial
dimensions.
RNN reuses the same recurrent block to process each single time frame in a
series and summarizes past time frames to feed into the next time frame.
The long-short term memory network (LSTM) is one of the most popular
implementations of RNN with an additional memory state for storing long-term
dependencies in sequences.
It can be coupled with CNNs to form a convolutional LSTM
(ConvLSTM)~\cite{xingjian2015convolutional}, replacing all matrix
multiplications with convolution operations to simultaneously model spatial 
and temporal aspects of data.
Transformer relies on an attention mechanism to determine dependency between each
input and output time frame, calculating a similarity score between each input
time frame (a query) and a set of context (keys) to serve as importance
weighting for a linear combination of the values that correspond to the
individual keys.
The query approach decouples the similarity and information aspect as the keys
and values---in contrary to past attention mechanisms---can now be different.
It also allows the output at each time frame to pull information from any
(masked) input time frame, thus excelling in modeling sequential dependency.
However, Transformers are huge architectures requiring large amounts of data to
train, 
which exceeded the resources of the authors at the time this research was conducted.

\begin{figure}
\includegraphics[width=\textwidth]{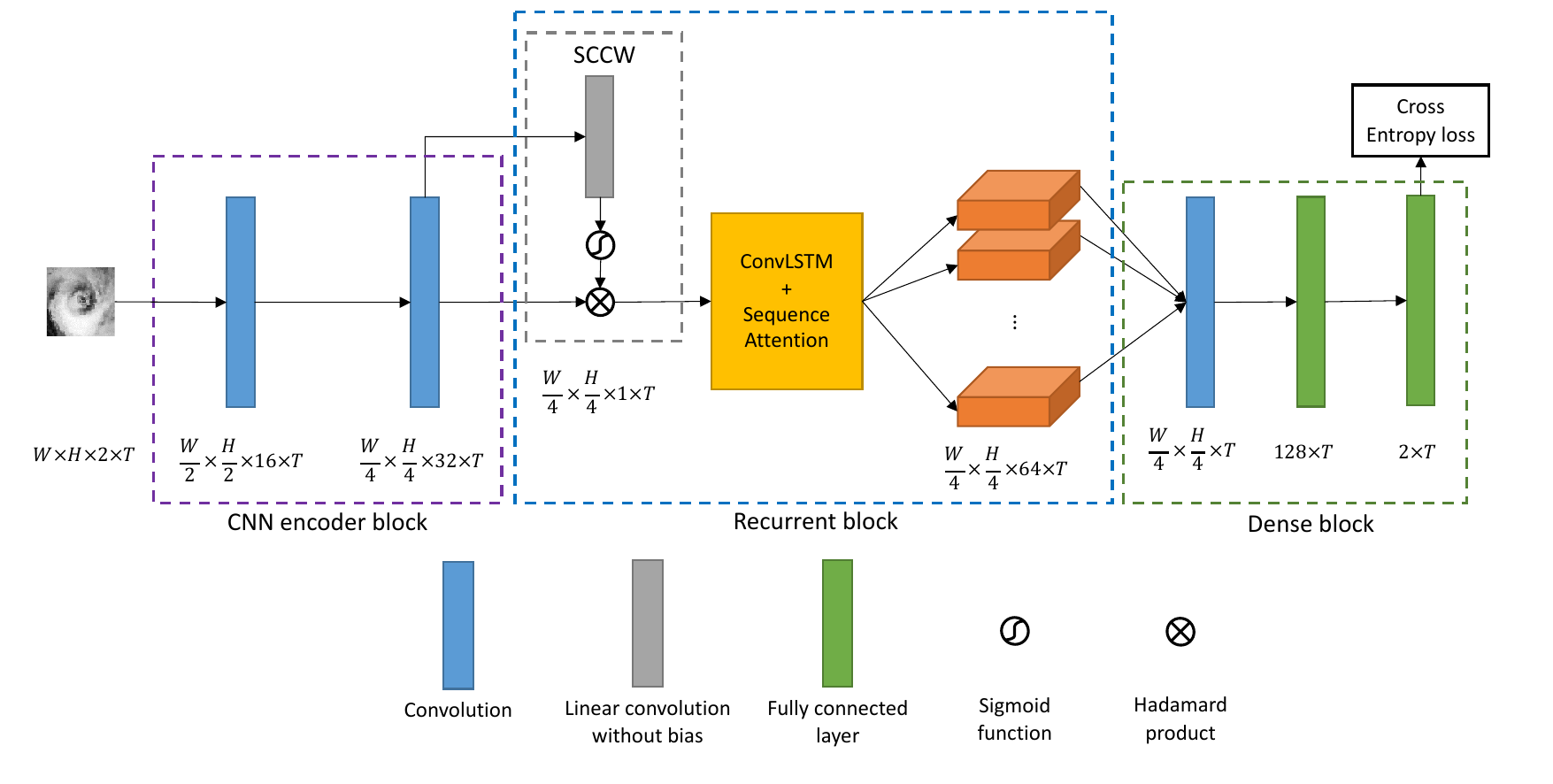}
\caption{Model architecture. The proposed model is split into three main
components: CNN encoder block, recurrent block, and dense block.}
\label{fig:architecture}
\end{figure}

Most pretrained computer vision neural networks are pretrained on
large-scale image classification datasets such as CIFAR-10 or ImageNet.
As the domain for satellite images is fundamentally different from those represented in
publicly-available pretrained models, it is difficult to apply common image
models (for instance ResNet50~\cite{he2016deep}) to this task.
Due to the lack of data, training from scratch was not feasible for larger models (as verified empirically).
In the end, we opted for the minimalist model architecture presented in
Fig.~\ref{fig:architecture}.
The base structure is a ConvLSTM model with convolutional layers in the
front to enrich features and a dense block in the end to compress the
three-dimensional features to one single probabilistic scalar output.

\subsection{Incorporating Meteorological Knowledge}
Neural network models in deep learning are notorious for their lack of
explainability and interpretability.
This may work without major drawbacks in common tasks such as image
classification or language translation, as the purpose of these tasks is to
autonomize previously human-labor-intensive jobs.
The relatively lower cost of making mistakes in these tasks allows for the application
of black-box models, as deep neural networks outperform past approaches.
However, for other tasks with more at stake (for instance in the medical domain),
being able to peek into the black box becomes significantly more important than
pure performance.
Understanding when and why a model makes mistakes helps people build trust 
the model output and evaluate the risk of incorporating it into the workflow.
For instance, saliency maps~\cite{simonyan2013deep} are utilized to trace which
part of the image contributes most to the classification result.

One of the most intuitive methods of understanding models is to directly design
the model according to previously acquired domain knowledge.
By exploiting such know-how, the learning is guided by expert insight which
allows analysis of whether the models learn as intended.
As mentioned in earlier sections, the Advanced Dvorak Technique used for TC
intensity estimation inspired the attention components of our model.
The major purpose of these additional components is to help diagnose and
understand the model, as opposed to improving its performance.

\subsubsection{Cross Channel Attention (CCA)}
The first of the two key points of ADT is to locate the TC eye and perform
scene-type analysis.
The extreme convection in the core area of TCs is known to be critical for RI.
We integrate this into our model by applying cross channel attention
between the CNN encoder block and recurrent block to reinforce the key
locations for the model to focus on.
Specifically, cross channel attention takes the multiple channel feature and apply a 2-dimensional importance weighting mask that is shared across every channel.
In our model, the importance mask is implemented as a two-layer CNN with sigmoid function in the end for normalization.
The global view across all different channels allows incorporation of
information from different sources to facilitate an overall importance evaluation of
where in the hidden map the model should focus its attention.
Given our domain knowledge, we expect a successful model to focus mainly on the TC core.

\subsubsection{Sequence Self-attention (SSA)}
The second key point of ADT is to apply time averaging on the predicted
intensity, showcasing the importance of taking earlier information into
account.
In ADT the averaging scheme calculates the non-weighted mean intensity over a
3-hr window (empirically determined).
Directly translating the concept to RI prediction is problematic, as
(1) our model predicts the occurrence of a future event, so the
ground truth for the recent past is still not available; and (2) averaging the
RI probability may not be as reasonable as averaging TC intensity.
Also, we are not satisfied with the heuristic of applying the non-weighted average
over a fixed window as, intuitively, the optimal contribution of different time
steps is certainly different and should be determined dynamically.
As a result, we incorporate the widely adopted sequence attention
mechanism~\cite{luong2015effective} into our recurrent block.
Of the different styles of sequence attention, we base our design on 
Bahdanau style self-attention~\cite{bahdanau2014neural}.
For each time step, features with information for the current time step is concatenated with a weighted average of features from all time steps prior to form the recurrent block input.
The weighting is calculated by evaluating the similarity between features for the current time step and prior features.
In our model, we use the projected cosine distance as the similarity metric, which is implemented as first passing the two given features through a two-layer CNN for feature extraction, then applying normalized inner product.
The CNN for feature extraction before calculating the cosine distance is essential due to the self-rotation motion of tropical cyclones.
Pixel misalignment caused by rotation is remedied by adding the additional feature extractor, which implicitly learns to transform the features into higher level ones with rotation-invariance.

\section{Experiment and Analysis}
\label{sec:experiment_and_analysis}
For the experiments, we trained the model on the training set for 500
epochs, validating on the validation set every 10 epochs.
Hyper-parameter tuning and epoch selection were achieved by selecting the
model with best validation PR-AUC.
We then evaluated the trained model on the test set and report the test
performance in this paper.
Hyper-parameters were minimally tuned only for the base ConvLSTM model such that the model converges and does not
collapse to predicting the majority class only; for all other variants,
they were kept identical.
Tuned parameters include
\begin{itemize}
    \item Batch size: 256
    \item Learning rate: 5e-4
    \item L2 regularization factor: 1e-5
    \item Dense layer dropout rate: 0.2.
\end{itemize}
All weights are initialized identically according to convention except for the final logits layer bias,
which was carefully set to $\log(\frac{\text{\# of positive class instances}}{\text{\# of
negative class instances}})$ to output a closer guess initially before any training given class imbalance.
We chose the Adam optimizer~\cite{kingma2014adam} to optimize the weighted binary
cross entropy loss with class weights corresponding to the reciprocal of the number
of class instances ($1:20$).
We experimented with both class weighting and minor class bootstrapping to prevent
model collapse and found that the latter caused the model to overfit.
The batch size was set as large as possible to mitigate large variances in
gradient magnitude when the number of sampled instances was different for each
mini-batch.
The input satellite images were augmented by random rotation to simulate the
natural rotation of tropical cyclones.
This is the key to making deep learning possible, as without rotation
augmentation, even the most shallow neural network model easily overfits the
training set.
All the code and trained models are publicly 
released.\footnote{\url{https://github.com/DixonCider/TCRI-Benchmark}}

\subsection{Input Time Length Experiment}
\begin{figure}
\centering
\begin{subfigure}{.455\textwidth}
  \centering
  \includegraphics[width=\linewidth]{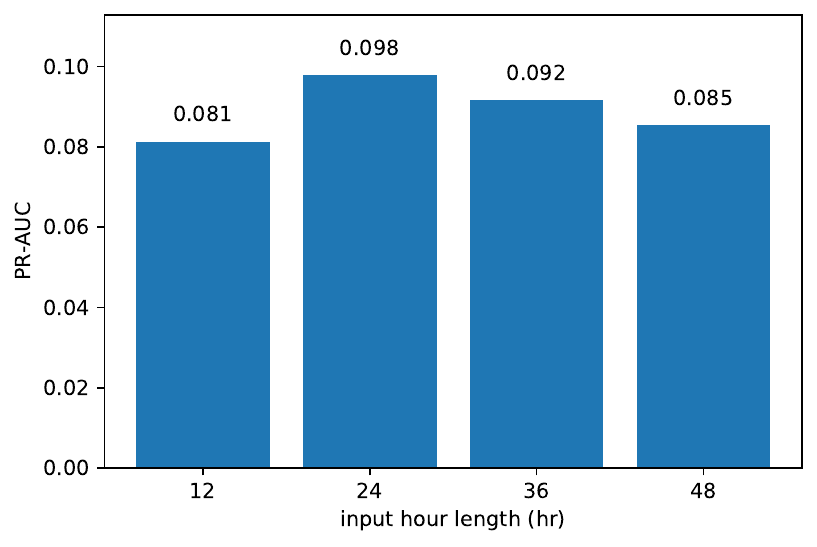}  
  \label{fig:diff_input_len_prauc}
\end{subfigure}
\begin{subfigure}{.455\textwidth}
  \centering
  \includegraphics[width=\linewidth]{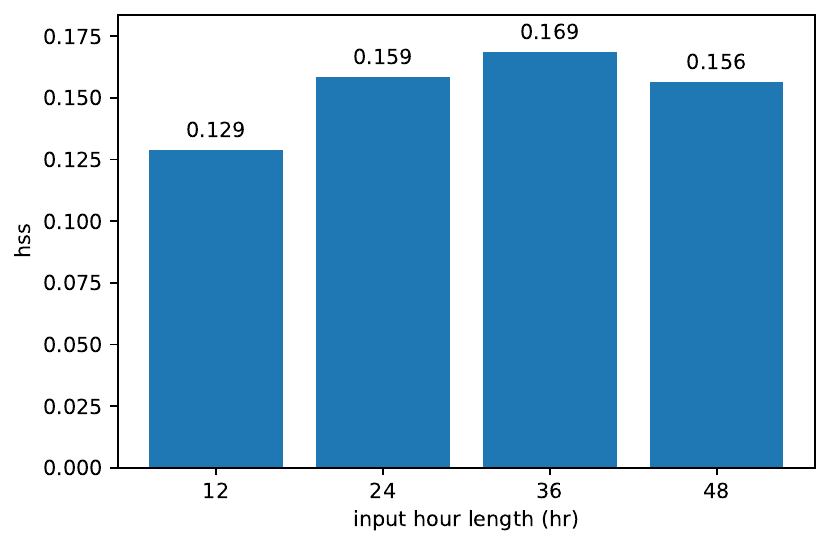}
  \label{fig:diff_input_len_hss}
\end{subfigure}
\caption{Performance given different input time lengths (left: PR-AUC, right: HSS).
These results show that input lengths of 24 or 36 are in
the range of optimal input time lengths.}
\label{fig:diff_input_len}
\end{figure}

The input time length is 
how long prior to the time step   
should be included as the input to the model.
It may seem that the longer the better, as the model can actively choose
what features to depend on.
However, tropical cyclones are fundamentally chaotic systems where small
changes in the initial value lead to largely bifurcated results.
In SHIPS-RII, 24 hours of data are fed into the model.
The trade-off for the input time length is that with a shorter length, the model
can start predicting in the very early stages of TC development but may use
more information from earlier stages, whereas with a longer length, the situation is the exact
opposite, and the training time is extended linearly with longer gradient
backpropagation.
As ConvLSTM can take arbitrary-length sequence input, we experimented
with dynamic- and fixed-length input and found that  dynamic-length resulted in inefficient
training (each sequence cannot be stacked uniformly)
and poor performance, suggesting that fixed-length input is better. 
We experimented with input lengths of 12, 24, 36, and 48 hours; the results
are presented in Fig.~\ref{fig:diff_input_len}.
For PR-AUC, an input length of 24 hours yields the best performance, identical to
the empirically determined length adopted in past studies. 
An input length of 36 hours performs satisfactorily as well, indicating the possibility of
the model benefiting from more data.
For HSS, an input length of 36 hours yields the best performance, in contrast to
the empirically determined length in past studies, with 24 and 48
hours not far behind.
Thus, input lengths of 24 or 36 are in the range of optimal input time lengths.
Interestingly, this alignment with meteorology-domain knowledge suggests that
approaches from past studies and our model, as different as they might be,
find important features from a similar range in history.
For the sake of faster training and less data required for prediction, we
conducted the following experiments with 24 hours as input.

\subsection{Ablation Study}
We conducted an ablation study on the two ADT-inspired components attached to the
base ConvLSTM model.
Recall that the purpose of CCA and SSA is to assist in the understanding of the model
prediction with gains in model performance as a secondary goal.
As a result, the hyperparameters are not specifically tuned for each model, nor
is the architecture.
Performance gains via optimizing these details is left as future work.
The PR-AUC and HSS results of the different variants are shown in Table~\ref{tab:performance}.

ConvLSTM, ConvLSTM + CCA, and ConvLSTM + SSA perform similarly in both
metrics, which shows that the additional domain-inspired components do not
harm prediction quality.
Note however that the combination of CCA and SSA results in
clearly worse performance as the model size becomes too large,
causing the model to overfit on the training set.
This is an example of the most difficult issue we encountered when designing
the model: balancing having enough capacity to model spatiotemporal relations
and being compact enough to not overfit on the small dataset.
We will discuss the takeaways from CCA and SSA in the following sections.

\begin{table}
\caption{Results of ablation study}\label{tab:performance}
\centering
\begin{tabular}{|l|l|l|}
\hline
                                & PR-AUC    & Heidke Skill Score\\
\hline
ConvLSTM                        & 0.098     & 0.159\\
ConvLSTM + CCA                & 0.095     & 0.164\\
ConvLSTM + SSA              & 0.099     & 0.161\\
ConvLSTM + CCA + SSA       & 0.089     & 0.152\\
\hline
\end{tabular}
\end{table}

\subsection{Cross Channel Attention Analysis}
\begin{figure}
\centering
\begin{subfigure}{.455\textwidth}
  \centering
  \includegraphics[width=\linewidth]{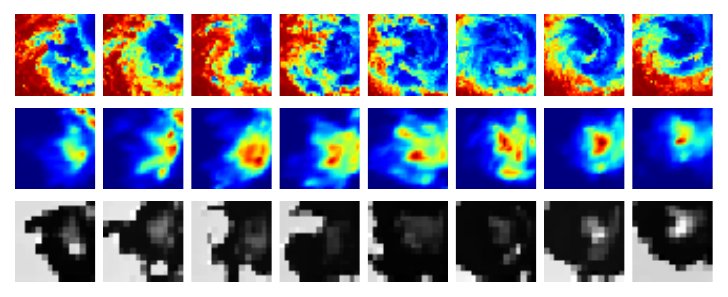}  
\end{subfigure}
\begin{subfigure}{.455\textwidth}
  \centering
  \includegraphics[width=\linewidth]{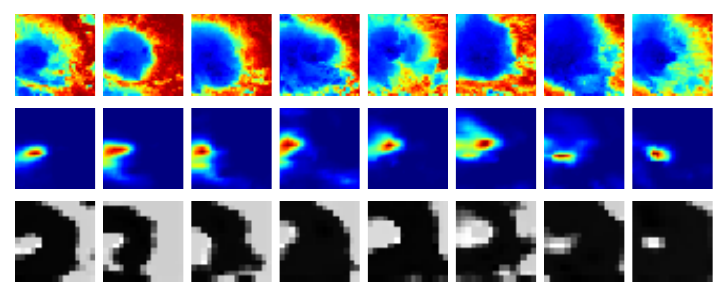}
\end{subfigure}
\begin{subfigure}{.455\textwidth}
  \centering
  \includegraphics[width=\linewidth]{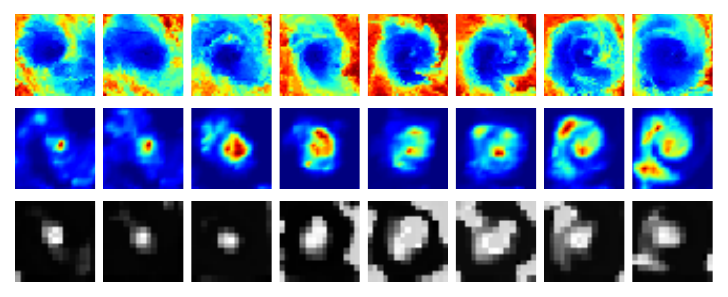}  
\end{subfigure}
\begin{subfigure}{.455\textwidth}
  \centering
  \includegraphics[width=\linewidth]{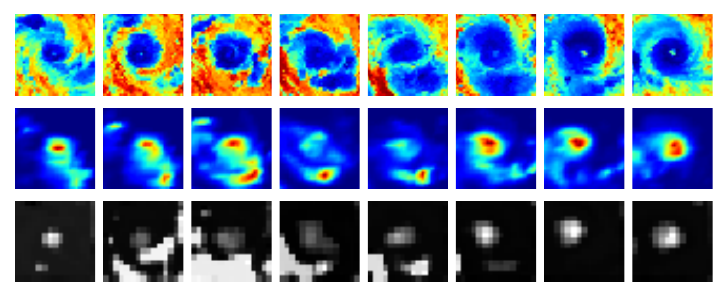}
\end{subfigure}
\caption{Cross channel attention. 
Four randomly sampled instances are visualized, each with three rows; the top, middle,
and bottom being the satellite IR1 channel, satellite PMW channel, and model
attention map (CCA), respectively. 
White implies a higher importance for the weighting.}
\label{fig:sccw_visualize}
\end{figure}

The purpose of CCA is to combine information across all channels to produce a shared importance mapping.
In Fig.~\ref{fig:sccw_visualize} we randomly sampled four instances for
visualization and compare these with existing domain knowledge in meteorology
regarding factors relevant to rapid intensification.
We observe that CCA helps the neural network to focus on coherent convective
features within the TC core region.
High importance is given to areas with extreme convection characterized by
high PMW rain rates (warm color in the figures) and extremely low IR1
brightness temperatures (dark blue in the figures), indicating that cross
channel information is jointly considered.
The strongly subsidence and dry area in the TC outer region characterized by
zero PMW rain rate and very high IR1 brightness temperatures are also
highlighted by the weighting.
Domain knowledge holds that RI is closely related to the convective burst
within the TC inner core.
Therefore, we conclude that the focus of the model from a meteorological perspective
is crucial to making correct predictions.

Furthermore, we demonstrate how the model follows the dynamical TC development
process with the bottom-right subfigure in Fig.~\ref{fig:sccw_visualize}.
We observe the formation of the TC eye (high IR1 Tb at the center)
indicating the potential intensification of the TC in the last three IR1
satellite images.
Importance weighting becomes significantly stronger at the TC center,
focusing on the evolution of TC inner-core convection.

\begin{figure}[h]
\centering
\includegraphics[width=.8\textwidth]{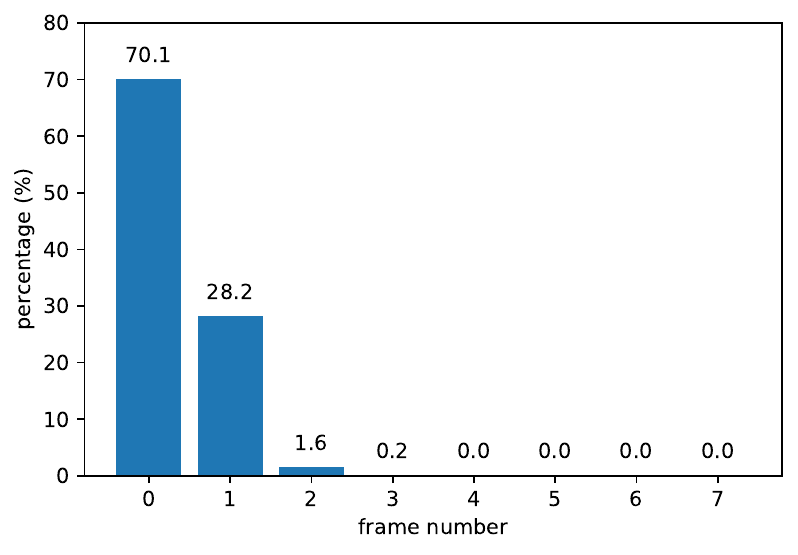}
\caption{Histogram of time step with highest attention score.
Counts are normalized with respect to total number of instances, presented here as percentages.}
\label{fig:seqattn_hist}
\end{figure}

\subsection{Sequence Self-attention Analysis}
In ADT, an intensity average is taken over a 3-hr window with equal weighting,
which is counter-intuitive as the importance of each time step should be
different.
We inspected the attention weights of the last time step over the time series
extracted from the sequence attention component.
We assume that since later time steps are closer to the
final step, their similarities are higher, analogous to the exponential
weighted moving average giving higher weights to more recent information.
In Fig.~\ref{fig:seqattn_hist} we present the histogram of the time step 
associated with the highest attention score.

Interestingly, the attention is highest on the first frame and exponentially
decays over time.
The last four steps have never been given the highest attention score out of
all time steps, going against our intuition that later frames are more
important.
This phenomenon is explained by a key difference between our approach toward
the weighted average versus the time averaging in ADT.
In ADT, as the averaging is applied directly to intensity,
due to the changes in atmospheric environment being continuous, intensity
remains similar between time steps that are closer.
In our model, the weighted average of sequence attention is applied to the
hidden maps derived from the input satellite image features.
Neighboring time steps provide similar or even redundant information whereas more 
distant time steps offer more.
As the entire 24-hr duration is important, the last frame gives more attention
to the distant past to retrieve features that may not be sufficiently encoded
in the ConvLSTM memory.
It is also possible that since the end goal is to predict occurrences of rapid
intensification, which is defined as whether the wind speed of the last frame increases by over
35 knots within 24 hours, the model looks back in time to determine the
difference between features from now and 24 hours ago as a point of reference.

\section{Conclusion}
\label{sec:conclusion}
In this work, we propose a new benchmark for detecting rapid intensification of 
tropical cyclones, including a benchmark dataset based on TC satellite
images with clearly defined, disjoint training, validation, and testing sets.
We seek to exploit features from lower-level data such as satellite signals
as they offer the most detailed information, in contrast to features
utilized in the past derived from trial-and-error heuristics with a mix of
measured and simulated data.
We point out the insuitability of the commonly-reported Brier skill score 
given the nature of the high class imbalance of the task, and replace it with
the precision-recall area under the curve (PR-AUC) and the Heidke skill score (HSS) to
better reflect true model prediction performance.
We also explore the potential of applying deep learning methods, which 
are known to benefit from large quantities of detailed spatiotemporal
data which humans are unable to directly process.
The proposed model is based on the ConvLSTM network with additional 
components---cross channel attention (CCA) and sequence self-attention 
(SSA)---inspired by the Advanced Dvorak Technique, a meteorological technique used to
estimate TC intensity.
We examine importance weighting from CCA and observe that the model 
focuses on the TC core where extreme convection occurs, a key factor for rapid
intensification.
We also evaluate sequence self-attention to determine which time step in a 
24-hr series is paid the most attention by the model.
The overall results suggest that important prediction features are spread
across the entire series.
In future work, we will better incorporate CCA and SSA to maintain
model performance and hopefully incorporate all channels of satellite data.

%
%

\section*{Acknowledgements}
The project was partially supported by the Ministry of Science and Technology in Taiwan via MOST 107-2628-E-002-008-MY3 and 108-2119-M-007-010.


\end{document}